\documentclass{article} 
\usepackage{iclr2020_conference,times}

\usepackage{amsmath,amsfonts,bm}

\def\eqref#1{equation~\ref{#1}}

\def\1{\bm{1}}

\DeclareMathAlphabet{\mathsfit}{\encodingdefault}{\sfdefault}{m}{sl}
\SetMathAlphabet{\mathsfit}{bold}{\encodingdefault}{\sfdefault}{bx}{n}

\usepackage{hyperref}
\usepackage{url}

\usepackage{subfigure}
\usepackage{multirow}
\usepackage{booktabs}
\usepackage{multicol}
\usepackage{verbatim}

\usepackage{algorithm}
\usepackage{algorithmicx}  
\usepackage{algpseudocode}
\usepackage{bm}
\usepackage{amssymb}
\usepackage[misc]{ifsym}
\usepackage{makecell}
\usepackage{appendix}
\usepackage{graphicx}

\newcommand{\orw}[1]{\textcolor{green}{\textbf{#1}}}
\newcommand{\mow}[1]{\textcolor{red}{\textbf{#1}}}

\newcolumntype{L}[1]{>{\raggedright\arraybackslash}p{#1}}  
\newcolumntype{C}[1]{>{\centering\arraybackslash}p{#1}}  
\newcolumntype{R}[1]{>{\raggedleft\arraybackslash}p{#1}}

\newcommand{\TheName}{ FastWordBug }

\title{FastWordBug: A Fast Method To Generate Adversarial Text Against NLP Applications}

\iclrfinalcopy

\author{Dou Goodman\textsuperscript{\Letter} , Lv Zhonghou \& Wang Minghua\\
Baidu X-Lab\\
Beijing, China\\
\texttt{liu.yan@baidu.com}
}

\begin{document}

\maketitle

\begin{abstract}
In this paper, we present a novel algorithm, FastWordBug, to efficiently generate small text perturbations in a black-box setting that forces a sentiment analysis or text classification mode to make an incorrect prediction. By combining the part of speech attributes of words, we propose a scoring method that can quickly identify important words that affect text classification. We evaluate FastWordBug on three real-world text datasets and two state-of-the-art machine learning models under black-box setting. The results show that our method can significantly reduce the accuracy of the model, and at the same time, we can call the model as little as possible, with the highest attack efficiency. We also attack two popular real-world cloud services of NLP, and the results show that our method works as well.
\end{abstract}

\section{Introduction}
As far as we know, \citet{Papernot2016Crafting} first studied the problem of adversarial example in text. \citet{Papernot2016Crafting} contributed to the field of adversarial machine learning by investigating adversarial input sequences for recurrent neural networks processing sequential data and they showed that adversaries can craft adversarial sequences misleading both categorical and sequential recurrent neural networks.


Although various techniques have been proposed to generate adversarial example for white-box attacks on text\citep{Papernot2016Crafting,Ebrahimi2017HotFlip,Samanta2017Towards,Sato2018Interpretable,gong2018adversarial,Liang2018Deep}, little attention has been paid to black-box attacks, which are more realistic scenarios. 

In the case of black-box attack, the most important step is to find the most important words that affect the result of the model. \citet{Gao2018Black} proposed DeepWordBug, they employed novel scoring strategies to identify the critical words that, if modified, cause the classifier to make an incorrect prediction. \citet{Li2018TextBugger} proposed TEXTBUGGER, a framework that can effectively and efficiently generate utility-preserving adversarial texts against state-of-the-art text classification systems under both white-box and black-box settings. In the black-box scenario, \citeauthor{Li2018TextBugger} first found the important sentences, and then used a scoring function to find important words to manipulate.


\citet{Gao2018Black,Li2018TextBugger} call the model and measure the importance of each word according to the change of the returned confidence value. We have studied the relationship between the POS tag (part-of-speech tag) of words and the importance of words influencing the classification results of the model. The experimental results show that by combining the part of speech of words, we can greatly reduce the number of calls to the model or cloud service of NLP and improve the efficiency of generating adversarial text, while the attack is still effective. A POS tag is a special label assigned to each token (word) in a text corpus to indicate the part of speech and often also other grammatical categories such as tense, number (plural/singular), case etc. POS tags are used in corpus searches and in text analysis tools and algorithms\footnote{https://www.sketchengine.eu/pos-tags/}.

\citet{belinkov2017synthetic,Alzantot_2018,Eger_2019} can be regarded as innovative word modification methods, which can be integrated with our algorithm.



\section{Attack Design}
\subsection{Problem Formulation}
The function of a pre-trained classification model $F$, e.g. a sentiment analysis or a text classification model, is mapping from input set to the label set. For a clean text example $O$, it is correctly classified by $F$ to ground truth label $y \in Y$, where $Y$ including $\{1,2, \ldots, k\}$ is a label set of $k$ classes. An attacker aims at adding small perturbations in $O$ to generate adversarial example $ADV$, so that $F(ADV) \neq F(O)$, where $D(ADV,O)<\epsilon$. $D$ captures the semantic similarity between $ADV$ and $O$, $\epsilon$ is a threshold to limit the size of perturbations.
\subsection{Threat Model}
We assume the attacker has black-box access to the target model: the attacker is not aware of the model architecture, parameters, or training data, and is only capable of querying the target model with supplied inputs and obtaining the output predictions and their confidence scores. We chose to use untargeted attack i.e., changing the model’s output, because it is more suitable as a benchmark method.

\begin{small}
\begin{equation} \label{eq:word_contribution}
C_{w_j}\!=\! \mathcal{F}_y(w_1, w_2, \cdots \!, w_m) -\! \mathcal{F}_y(w_1, \cdots \!, w_{j\!-\! 1}, w_{j\!+\! 1}, \cdots \!, w_m)
\end{equation}
\end{small}

\begin{algorithm}[tp] 
\caption{Calculate the weight of different POS tags}  
 \label{alg:pos_tag}
\begin{algorithmic}[1] 
\Require legitimate documents $D$, classifier $\mathcal{F(\cdot)}$
\Ensure The weight of different POS tags $W(t)$
\State $t$ is a POS tag, $R\{t\}$ records the number of times $t$ has the maximum confidence.
\State Inititialize hashtable $R$,  

\For{$(x,y)$ in $D$}  
    \For {$w_i$ in $x$}
	    \State Compute $C_{w_i}$ according to Eq.\ref{eq:word_contribution};\
	\EndFor
    \State $W_{ordered} \gets Sort(words)$ according to $C_{w_i}$;\
	\State Get the word $w_j$ which has the maximum confidence maximum reduction $C_{w_i}$;\
	\State Get the POS tag $t$ of $w_j$;\
	\State $R(t) \gets R(t)+1$;\
\EndFor
\State $W(t)=softmax(R(t),R(\cdot))$;\
\State \Return $W(\cdot)$. 
\end{algorithmic}  
\end{algorithm}

\subsection{\TheName}
 We propose Algorithm \ref{alg:pos_tag} to quantify the relationship between the POS tags of words and the importance of words influencing the classification results of the model. We suppose that text $x$ is split into a series of words: $w=\{w_1,w_2\,...,w_n\}$. The importance score of $w$ is represented with the confidence value of the predicted class $\mathcal{F}_y(x)=\mathcal{F}_y(w)$. Following \citet{Gao2018Black,Li2018TextBugger}, the importance of the $j^{th}$ word in $w$ is:

We evaluate the weight of a POS tag as the performance reduction when the corresponding word is disabled in the text. Intuitively, removing a word with greater capacity of discrimination usually causes higher confidence reduction. In this way, such word or say POS tag should be pay more attention .i.e more weight. In addition, we use the $softmax$\footnote{https://en.wikipedia.org/wiki/Softmax\_function} function to convert the statistical weight value on the specified data set to between $0$ and $1$, which is easy to use.

We propose \TheName to generate word-based adversarial text under black box setting. 

\textbf{Step 1: Find Important Sentences.} We divide the text into a sequence of sentences. Then, by calling the model, the confidence of each sentence is calculated in turn. Finally, by sorting the confidence values, a sentence sequence sorted by importance is obtained. The first sentence means more important, but it will also be prioritized.

\textbf{Step 2: Filter Important Words with POS tag weight.} We divide the ordered sentence sequence into words and mark the part of speech. Then we filter the words whose POS tag weight exceeds the threshold and return them in order. In this process, we do not need to call the model, but make full use of POS tag features of words. 

\textbf{Step 3: Modify Important Words.} There are many ways to modify words. Following \citet{Gao2018Black}, we use four basic methods:
\begin{itemize}
    \item Swap: Swap two adjacent letters in the word.
    \item Substitution: Substitute a letter in the word with a random letter.
    \item Deletion: Delete a random letter from the word.
    \item Insertion: Insert a random letter in the word.
\end{itemize}

All four methods are designed to produce a word that is not in the model's dictionary. We modify the filtered words in turn, and use these four methods for each word in turn. If the modification of the word can lead to a decrease in the confidence of the model for the original label, the modification for the word will be retained. If several modifications can decrease the confidence, we choose the most effective. In this step, the model is called four times for each word. When the classification label changes or the difference between the adversarial text and the original text exceeds the threshold, the loop exits ahead of time.


\section{Attack Evaluation}
\subsection{Datasets}
\textbf{Internet Movie Database (IMDB)\footnote{http://ai.stanford.edu/~amaas/data/sentiment/}}: It is crawled
from Internet including 50000 positive and negative reviews and average length of the review is nearly 200 words. It is usually used for binary sentiment classification including richer data than other similar datasets\citep{maas-EtAl:2011:ACL-HLT2011}.


\textbf{AG's News\footnote{http://www.di.unipi.it/~gulli/AG\_corpus\_of\_news\_articles.html}}: It is constructed by choosing 4 largest classes from the original corpus. Each class contains 30,000 training samples and 1,900 testing samples. The total number of training samples is 120,000 and testing 7,600\citep{Xiang2015Character}.


\subsection{Targeted Models}
To show that our method is effective, we perform our experiments on two well trained models: \textbf{Word-LSTM\citep{Gao2018Black}} and \textbf{Text-CNN\citep{Kim_2014}}. Following \citet{Li2018TextBugger}, all models are trained in a hold-out test strategy, and hyper-parameters were tuned only on the validation set. We launch a black box attack on these models.   
%

\subsection{Evaluation Metrics}
We use \textbf{Top-1 Accuracy} to measure the attack effect. For the same model, the lower the accuracy of classifying the adversarial text, the better the attack effect. The number of \textbf{Perturbed Word} is used to define the distance between the adversarial text and the original text. The number of calls of the model .i.e \textbf{\#Model Called } is used to measure the attack efficiency. The less the number of calls, the higher the attack efficiency.

\begin{table*}[t]
\caption{Results on the black box attack on IMDB datasets.}
\label{tab:main_result_imdb}
~\\
\centering
\resizebox{0.9\textwidth}{!}{

\begin{tabular}{lcccccc}
\toprule
{} & \multicolumn{2}{c}{Accuracy} & \multicolumn{2}{c}{\#Model Called} & \multicolumn{2}{c}{Perturbed Word} \\
 \cmidrule(lr){2-3} \cmidrule(lr){4-5} \cmidrule(lr){6-7} & Text-CNN & Word-LSTM & Text-CNN & Word-LSTM &       Text-CNN & Word-LSTM \\
\midrule
No Attack&87.0\% &      78.0\% &     N/A &      N/A  &           10.0\% &      10.0\% \\
DeepWordBug\citep{Gao2018Black} &     34.0\% &      48.5\% &   607.84 &    571.14 &           10.0\% &      10.0\% \\
TEXTBUGGER\citep{Li2018TextBugger}  &     16.5\% &      25.5\% &   551.50 &    570.15 &           10.0\% &      10.0\% \\
\TheName         &     \textbf{13.0\%} &      \textbf{21.0\%} &   \textbf{232.19} &    \textbf{273.12} &           10.0\% &      10.0\% \\
\bottomrule
\end{tabular}

}
\end{table*}

\begin{table*}[t]
\caption{Results on the black box attack on AG's News datasets.}
\label{tab:main_result_ag}
~\\
\centering
\resizebox{0.9\textwidth}{!}{

\begin{tabular}{lcccccc}
\toprule
{} & \multicolumn{2}{c}{Accuracy} & \multicolumn{2}{c}{\#Model Called} & \multicolumn{2}{c}{Perturbed Word} \\
 \cmidrule(lr){2-3} \cmidrule(lr){4-5} \cmidrule(lr){6-7} & Text-CNN & Word-LSTM & Text-CNN & Word-LSTM &       Text-CNN & Word-LSTM \\
\midrule
No Attack& 87.0\% &      88.0\% &      N/A &      N/A &           15.0\% &      15.0\%  \\
DeepWordBug\citep{Gao2018Black} &     51.5\% &      42.0\% &     92.71 &     90.97 &           15.0\% &      15.0\% \\
TEXTBUGGER\citep{Li2018TextBugger}  &     37.5\% &      35.0\% &     70.08 &     70.93 &           15.0\% &      15.0\% \\
\TheName         &     \textbf{36.0\%} &      \textbf{31.5\%} &     \textbf{62.02} &     \textbf{56.62} &           15.0\% &      15.0\% \\
\bottomrule
\end{tabular}

}
\end{table*}

\begin{table*}[!t]

\setlength\belowcaptionskip{-5pt}
\centering

\begin{tabular}{|p{12cm}|}
\hline
\textbf{Classifier}: Word-LSTM\\
\textbf{Original Text Prediction}: Sci/Tech (Confidence = 47.80\%)\\
\textbf{Adversarial Text Prediction}: Business (Confidence = 52.52\%)\\
\hline
\textbf{Original Text}: \textit{Tyrannosaurus rex achieved its massive size due to an enormous growth spurt \orw{during} its adolescent years.} \\
\textbf{Adversarial Text}: \textit{Tyrannosaurus rex achieved its massive size due to an enormous growth spurt \mow{durnig} its adolescent years.} \\
\hline
\hline

\textbf{Classifier}: Text-CNN\\
\textbf{Original Text Prediction}: Company (Confidence = 98.16\%)\\
\textbf{Adversarial Text Prediction}: Artist (Confidence = 20.27\%)\\
\hline
\textbf{Original Text}: \textit{Yates is a gardening company \orw{in} New Zealand and Australia.} \\
\textbf{Adversarial Text}: \textit{Yates is a gardening company \mow{i} New Zealand and Australia.} \\
\hline

\end{tabular}
\caption{Example of text classification task. Modified words are highlighted in green and red for the original and adversarial texts, respectively. Test samples are randomly selected from the test set of AG's News and DBPedia Ontology. }
\label{tab:demo_classifer}
\end{table*}

\subsection{Implementation}
We repeated each experiment 5 times and report the mean value. We randomly select 200 from the test set of IMDB and AG's News to generate the adversarial text, and select 200 to calculate the weight of POS Tag. The same data set uses the same weight value. In order to better measure the performance difference in selecting important words between different attack algorithms, we implement each algorithm to modify words in the same way as DeepWordBug\citep{Gao2018Black}. Words and sentences are splited by NTLK\footnote{http://www.nltk.org/}.
\subsection{Attack Performance}


The main experimental results are shown in Table \ref{tab:main_result_imdb} and Table \ref{tab:main_result_ag}, our attack algorithm leads in both efficiency and effect. In the sentiment analysis task on IMDB, our algorithm reduces the accuracy of Text-CNN and Word-LSTM models from 87.0\% and 78.0\% to 13.0\% and 21.0\% while \#Model Calls are also minimal. In the text classification task on AG’s News, our algorithm reduces the accuracy of Text-CNN and Word-LSTM models from 87.0\% and 88.0\% to 36.0\% and 31.5\% while \#Model Calls are also minimal. 

We also attack two popular real-world cloud services of NLP,.i.e \textbf{Aylien Sentiment}\footnote{https://developer.aylien.com/text\-api\-demo} and \textbf{ParallelDots}\footnote{https://www.paralleldots.com/text\-analysis\-apis\#sentiment}, the results show that our method works as well, Table~\ref{tab:demo_aylien} and Table~\ref{tab:demo_paralleldots} are examples.

\begin{table*}[!t]
\setlength\belowcaptionskip{-5pt}
\centering
\begin{tabular}{|p{12cm}|}
\hline
\textbf{Original Text Prediction}: Negative (Confidence = 61 \%) \\
\hline
\textit{This is \orw{not} an in-depth review, \orw{but} FMLB neither deserves \orw{nor} requires one. You might enjoy it if you're a fan of \orw{bad} movies.} \\
\hline
\hline
\textbf{Adversarial Text Prediction}: Positive (Confidence = 68\%)\\
\hline
\textit{This is \mow{nt} an in-depth review, \mow{bt} FMLB neither deserves \mow{nr} requires one. You might enjoy it if you're a fan of \mow{bd} movies.} \\
\hline
\end{tabular}
\caption{Example of \TheName for the sentiment analysis task of \textbf{Aylien Sentiment}. Modified words are highlighted in green and red for the original and adversarial texts, respectively. Test data is randomly selected from the test set of IMDB. Space limited, we only show modified sentences. }
\label{tab:demo_aylien}
\end{table*}
\begin{table*}[!t]
\setlength\belowcaptionskip{-5pt}
\centering
\begin{tabular}{|p{12cm}|}
\hline
\textbf{Original Text Prediction}: Negative (Confidence = 68.90 \%) \\
\hline
\textit{This movie has the feel of a college project over it, who wants to do a blair witch project meets saw theme. But it \orw{isn't} successful. The cinematography is poor, and the acting even more so. The characters, in my opinion doesn't come off as being credible at all. The editing of the film isn't really working as intended either. There are a lot of poor effects, \orw{which} I believe are put in there to try and add a horrid effect. But to me it just gives me a feeling of indifference. I would stay away from this movie, unless you are a dedicated movie freak, who likes to watch "different" and indie "horror" movies. However, I believe this movie is not worth watching, for the average person. You will get no pleasure out of the poor effects, and the handycam feel, which this movie bestows on it's viewers.} \\
\hline
\hline
\textbf{Adversarial Text Prediction}: Positive (Confidence = 54.40\%)\\
\hline
\textit{This movie has the feel of a college project over it, who wants to do a blair witch project meets saw theme. But it \mow{is n't} successful. The cinematography is poor, and the acting even more so. The characters, in my opinion doesn't come off as being credible at all. The editing of the film isn't really working as intended either. There are a lot of poor effects, \mow{wihch} I believe are put in there to try and add a horrid effect. But to me it just gives me a feeling of indifference. I would stay away from this movie, unless you are a dedicated movie freak, who likes to watch "different" and indie "horror" movies. However, I believe this movie is not worth watching, for the average person. You will get no pleasure out of the poor effects, and the handycam feel, which this movie bestows on it's viewers.} \\
\hline
\end{tabular}
\caption{Example of \TheName for the sentiment analysis task of \textbf{ParallelDots}. Modified words are highlighted in green and red for the original and adversarial texts, respectively. Test data is randomly selected from the test set of IMDB.}
\label{tab:demo_paralleldots}
\end{table*}

\subsection{Transferability}
The result shown in Table \ref{tab:Transferability} demonstrates that the adversarial texts generated by \TheName can successfully transfer across multiple models. 

\begin{table}[h]
	\caption{Transferability on IMDB dataset. }
	\label{tab:Transferability}
	\centering
	\resizebox{0.45\textwidth}{!}{
	\begin{tabular}{cccc}
		\toprule
		Model & Text-CNN& Word-LSTM \\
		\midrule
		Text-CNN & 13.00\% & 64.50\%  \\
		Word-LSTM &  83.50\% & 21.00\% \\
		\bottomrule
	\end{tabular}
	}
\end{table}


\section{Conclusion}
Our main contributions can be summarized as follows.
\begin{itemize}
    \item We propose \TheName, a framework that can more effectively and efficiently generate adversarial text under black box setting.
    \item We propose an algorithm to quantify the relationship between the POS tags of words and the importance of words to generate adversarial text. 
    \item We evaluate \TheName on two real-world text datasets and two state-of-the-art machine learning models under black box setting. The results show that our method can significantly reduce the accuracy of the model, and at the same time, we can call the model as little as possible, with the highest attack efficiency.
\end{itemize}

\section*{Acknowledgments}
We would like to thank Jinfeng Li and Shouling Ji for their help and Shawn Ng for his code of Text-CNN\footnote{https://github.com/Shawn1993/cnn-text-classification-pytorch}, \citeauthor{Gao2018Black} for their code of DeepWordBug\footnote{https://github.com/QData/deepWordBug}. 

\bibliography{iclr2020_conference,public}

\begin{thebibliography}{14}
\providecommand{\natexlab}[1]{#1}
\providecommand{\url}[1]{\texttt{#1}}
\expandafter\ifx\csname urlstyle\endcsname\relax
  \providecommand{\doi}[1]{doi: #1}\else
  \providecommand{\doi}{doi: \begingroup \urlstyle{rm}\Url}\fi

\bibitem[Alzantot et~al.(2018)Alzantot, Sharma, Elgohary, Ho, Srivastava, and
  Chang]{Alzantot_2018}
Moustafa Alzantot, Yash Sharma, Ahmed Elgohary, Bo-Jhang Ho, Mani Srivastava,
  and Kai-Wei Chang.
\newblock Generating natural language adversarial examples.
\newblock \emph{Proceedings of the 2018 Conference on Empirical Methods in
  Natural Language Processing}, 2018.
\newblock \doi{10.18653/v1/d18-1316}.
\newblock URL \url{http://dx.doi.org/10.18653/v1/d18-1316}.

\bibitem[Belinkov \& Bisk(2017)Belinkov and Bisk]{belinkov2017synthetic}
Yonatan Belinkov and Yonatan Bisk.
\newblock Synthetic and natural noise both break neural machine translation,
  2017.

\bibitem[Ebrahimi et~al.(2017)Ebrahimi, Rao, Lowd, and
  Dou]{Ebrahimi2017HotFlip}
Javid Ebrahimi, Anyi Rao, Daniel Lowd, and Dejing Dou.
\newblock Hotflip: White-box adversarial examples for nlp.
\newblock 2017.

\bibitem[Eger et~al.(2019)Eger, Şahin, Rücklé, Lee, Schulz, Mesgar,
  Swarnkar, Simpson, and Gurevych]{Eger_2019}
Steffen Eger, Gözde~Gül Şahin, Andreas Rücklé, Ji-Ung Lee, Claudia Schulz,
  Mohsen Mesgar, Krishnkant Swarnkar, Edwin Simpson, and Iryna Gurevych.
\newblock Text processing like humans do: Visually attacking and shielding.
\newblock \emph{Proceedings of the 2019 Conference of the North}, 2019.
\newblock \doi{10.18653/v1/n19-1165}.
\newblock URL \url{http://dx.doi.org/10.18653/v1/n19-1165}.

\bibitem[Gao et~al.(2018)Gao, Lanchantin, Soffa, and Qi]{Gao2018Black}
Ji~Gao, Jack Lanchantin, Mary~Lou Soffa, and Yanjun Qi.
\newblock Black-box generation of adversarial text sequences to evade deep
  learning classifiers.
\newblock 2018.

\bibitem[Gong et~al.(2018)Gong, Wang, Li, Song, and Ku]{gong2018adversarial}
Zhitao Gong, Wenlu Wang, Bo~Li, Dawn Song, and Wei-Shinn Ku.
\newblock Adversarial texts with gradient methods, 2018.

\bibitem[Kim(2014)]{Kim_2014}
Yoon Kim.
\newblock Convolutional neural networks for sentence classification.
\newblock \emph{Proceedings of the 2014 Conference on Empirical Methods in
  Natural Language Processing (EMNLP)}, 2014.
\newblock \doi{10.3115/v1/d14-1181}.
\newblock URL \url{http://dx.doi.org/10.3115/v1/d14-1181}.

\bibitem[Li et~al.(2018)Li, Ji, Du, Li, and Wang]{Li2018TextBugger}
Jinfeng Li, Shouling Ji, Tianyu Du, Bo~Li, and Ting Wang.
\newblock Textbugger: Generating adversarial text against real-world
  applications.
\newblock 2018.

\bibitem[Liang et~al.(2018)Liang, Li, Su, Pan, and Shi]{Liang2018Deep}
Bin Liang, Hongcheng Li, Miaoqiang Su, Bian Pan, and Wenchang Shi.
\newblock Deep text classification can be fooled.
\newblock In \emph{IJCAI}, 2018.

\bibitem[Maas et~al.(2011)Maas, Daly, Pham, Huang, Ng, and
  Potts]{maas-EtAl:2011:ACL-HLT2011}
Andrew~L. Maas, Raymond~E. Daly, Peter~T. Pham, Dan Huang, Andrew~Y. Ng, and
  Christopher Potts.
\newblock Learning word vectors for sentiment analysis.
\newblock In \emph{Proceedings of the 49th Annual Meeting of the Association
  for Computational Linguistics: Human Language Technologies}, pp.\  142--150,
  Portland, Oregon, USA, June 2011. Association for Computational Linguistics.
\newblock URL \url{http://www.aclweb.org/anthology/P11-1015}.

\bibitem[Papernot et~al.(2016)Papernot, Mcdaniel, Swami, and
  Harang]{Papernot2016Crafting}
Nicolas Papernot, Patrick Mcdaniel, Ananthram Swami, and Richard Harang.
\newblock Crafting adversarial input sequences for recurrent neural networks.
\newblock 2016.

\bibitem[Samanta \& Mehta(2017)Samanta and Mehta]{Samanta2017Towards}
Suranjana Samanta and Sameep Mehta.
\newblock Towards crafting text adversarial samples.
\newblock 2017.

\bibitem[Sato et~al.(2018)Sato, Suzuki, Shindo, and
  Matsumoto]{Sato2018Interpretable}
Motoki Sato, Jun Suzuki, Hiroyuki Shindo, and Yuji Matsumoto.
\newblock Interpretable adversarial perturbation in input embedding space for
  text.
\newblock 2018.

\bibitem[Xiang et~al.(2015)Xiang, Zhao, and Lecun]{Xiang2015Character}
Zhang Xiang, Junbo Zhao, and Yann Lecun.
\newblock Character-level convolutional networks for text classification.
\newblock 2015.

\end{thebibliography}
\bibliographystyle{iclr2020_conference}

\newpage
\appendix
\section*{Appendix}

\section{Important tag of Part-of-speech tagging.}
\begin{table*}[htb]
\caption{Part of \textbf{Penn Treebank POS tags}. }
\label{tab:pos_tags}
~\\
\centering
\resizebox{0.9\textwidth}{!}{
\begin{tabular}{l|l|l}
\toprule
Tag &      Description & Example   \\
\midrule
 NN  &     noun, singular or mass  &  tiger, chair, laughter  \\
 NNP & noun, proper singular & Germany, God, Alice\\
DT & determiner	& the, a, these  \\
RB & adverb & extremely, loudly, hard\\
JJ & adjective & nice, easy\\
IN & conjunction, subordinating or preposition & of, on, before, unless\\
CC & conjunction, coordinating & and, or, but\\
VB & verb, base form & think\\
PRP & pronoun, personal & me, you, it\\
\bottomrule
\end{tabular}
}

\end{table*}

\end{document}